\documentclass[letterpaper,12pt]{article}
\usepackage{tabularx} 
\usepackage{amsmath}  
\usepackage{graphicx} 
\usepackage[margin=1in,letterpaper]{geometry} 
\usepackage{cite} 
\usepackage[final]{hyperref} 
\hypersetup{
	colorlinks=true,       
	linkcolor=blue,        
	citecolor=blue,        
	filecolor=magenta,     
	urlcolor=blue         
}
\usepackage{blindtext}
\usepackage[hashEnumerators,smartEllipses]{markdown}
\usepackage{xspace}
\usepackage{subcaption}
\usepackage[affil-it]{authblk} 
\usepackage{wrapfig}
\usepackage{booktabs}
\usepackage{multirow}
\usepackage{makecell}

\begin{document}

\title{Backdoors in DRL: Four Environments Focusing on In-distribution Triggers}
\author[1]{Chace Ashcraft\thanks{Chace.Ashcraft@jhuapl.edu}}
\author[1]{Ted Staley}
\author[1]{Josh Carney}
\author[1]{Cameron Hickert}
\author[2]{Derek Juba}
\author[1]{Kiran Karra\thanks{Contributions made while at Johns Hopkins University Applied Physics Laboratory}}
\author[1]{Nathan Drenkow}
\affil[1]{Johns Hopkins University Applied Physics Laboratory}
\affil[2]{National Institute of Science and Technology}
\maketitle

\vspace{-1.15cm}

\begin{abstract}
Backdoor attacks, or trojans, pose a security risk by concealing undesirable behavior in deep neural network models. Open-source neural networks are downloaded from the internet daily, possibly containing backdoors, and third-party model developers are common. To advance research on backdoor attack mitigation, we develop several trojans for deep reinforcement learning (DRL) agents. We focus on \textit{in-distribution} triggers, which occur within the agent's natural data distribution, since they pose a more significant security threat than \textit{out-of-distribution} triggers due to their ease of activation by the attacker during model deployment. We implement backdoor attacks in four reinforcement learning (RL) environments: \textit{LavaWorld}, \textit{Randomized LavaWorld}, \textit{Colorful Memory}, and \textit{Modified Safety Gymnasium}. We train various models, both clean and backdoored, to characterize these attacks. We find that in-distribution triggers can require additional effort to implement and be more challenging for models to learn, but are nevertheless viable threats in DRL even using basic data poisoning attacks. 
\end{abstract}

\vspace{-0.4cm}

\begin{figure}[h!]
    \def\bhp{0.8}
    \begin{subfigure}[t]{0.24\textwidth}
        \centering
        \includegraphics[height=\bhp\textwidth,width=\bhp\textwidth,trim=9 10 10 10,clip]{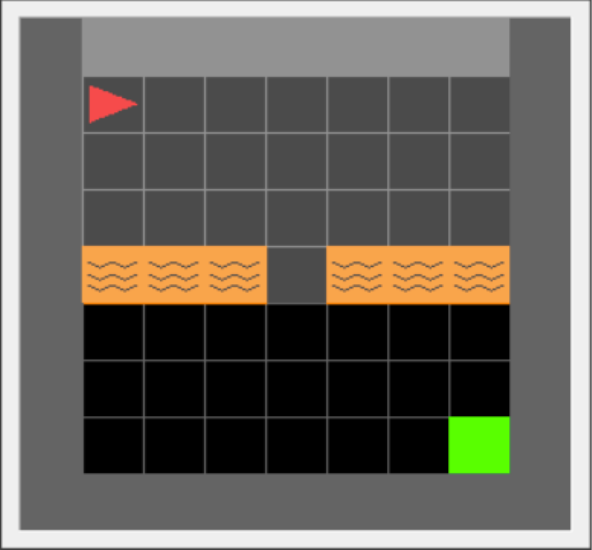}
        \caption*{LavaWorld}
        \label{fig:lavaworld_plain}
    \end{subfigure}
    \begin{subfigure}[t]{0.24\textwidth}
        \centering
        \includegraphics[height=\bhp\textwidth,width=\bhp\textwidth]{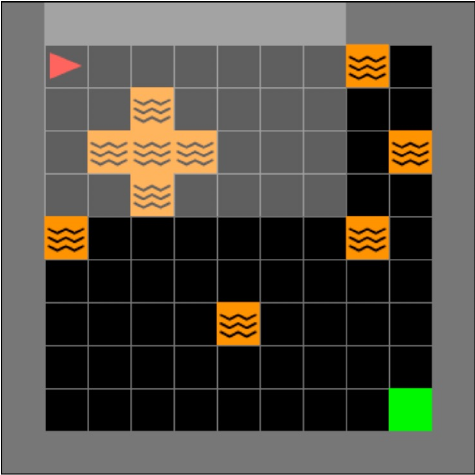}
        \caption*{\centering Randomized \break LavaWorld}
        \label{fig:rand_lavaworld_plain}
    \end{subfigure}
    \begin{subfigure}[t]{0.24\textwidth}
        \centering
        \includegraphics[height=\bhp\textwidth,width=\bhp\textwidth,trim=200 287 200 287, clip]{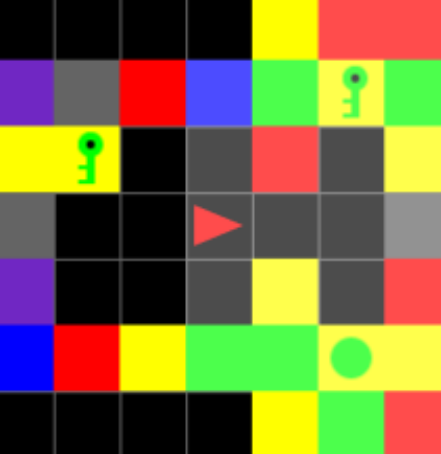}
        \caption*{Colorful Memory}
        \label{fig:cm_plain}
    \end{subfigure}
    \begin{subfigure}[t]{0.24\textwidth}
        \centering
        \includegraphics[height=\bhp\textwidth,width=\bhp\textwidth,trim=150 150 150 100, clip]{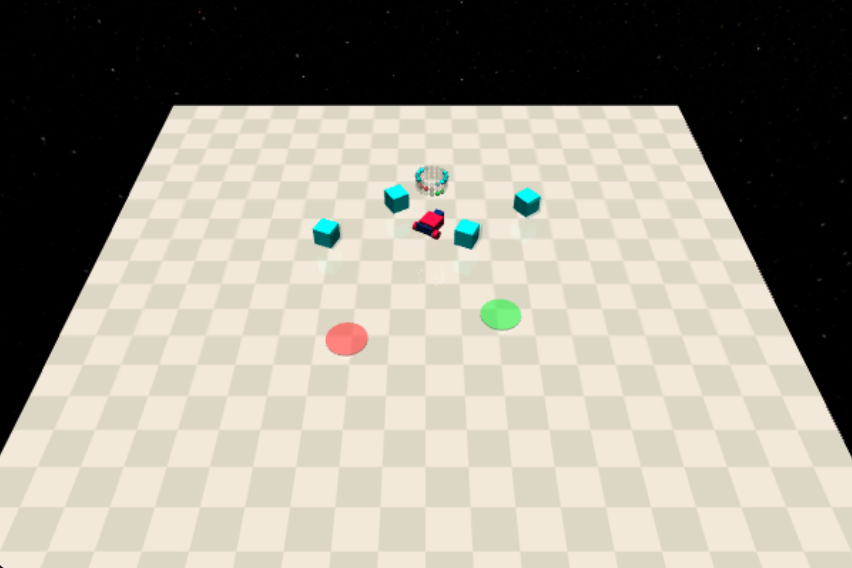}
        \caption*{\centering Modified \break Safety Gymnasium}
        \label{fig:mod_safety_gym}
    \end{subfigure}
    \caption*{Four environments for creating backdoored DRL agents.}
    \label{fig:envs_plain}
\end{figure}

\setcounter{figure}{0}

\pagebreak

\section{Introduction}
\label{sec:intro}

Deep learning is a powerful tool that has enabled new capabilities in machine perception and reasoning, but the black-box nature of its models may conceal unwanted behavior. Sometimes undesired behaviors are simply unintended consequences of low-quality data or poor training practices~\cite{rlblogpost, amodei2016concreteproblemsaisafety, alzubaidi2023survey}. However, they may also be intentionally introduced by malicious actors, through means such as backdoors or trojans~\cite{liu2018trojaning, gu2019badnetsidentifyingvulnerabilitiesmachine}. In backdoor attacks, the attacker, who has access to the model at training time, adjusts the data and training algorithm so that the model learns to perform the expected task until exposed to a trigger signal specified by the attacker. When a victim uses the model, it will appear to behave normally until the specific signal is provided, at which point the model will perform whatever behavior the attacker chose to inject into the model. To facilitate investigation and mitigation of these phenomena, we adapt backdoor techniques from the literature to the deep reinforcement learning (DRL) domain to be used for further test and evaluation. 

Three of our environments, \textit{LavaWorld}, \textit{Randomized LavaWorld}, and \textit{Colorful Memory}, are based on the popular MiniGrid~\cite{MinigridMiniworld23} RL environment library, which contains many grid-based navigation tasks of varying difficultly. Modifying environments allowed us to develop and study in-distribution triggers~\cite{ashcraft2021poisoningdeepreinforcementlearning}, which are triggers that occur naturally within the data distribution observed by the model during training, evaluation, and deployment. In contrast, out-of-distribution triggers consist of valid model inputs but fall outside of those distributions, requiring trigger insertion to occur between digitization of a signal by a sensor and the signal being given to the neural network. Thus, in-distribution triggers are easier to activate by an attacker than out-of-distribution triggers, and pose a more significant threat to neural networks. More discussion on the distinction between in-distribution and out-of-distribution triggers may be found in Appendix~\ref{subsec:indist-outdist}. Our fourth environment, \textit{Modified Safety Gymnasium}, is based on the popular Safety Gymnasium~\cite{ji2023safety} environment. We create an in-distribution trigger for this environment that allows us to explore more realistic robotic control tasks with continuous action spaces. In this report, we describe each environment in detail, explain how we trained our own backdoored models using data poisoning~\cite{gu2019badnetsidentifyingvulnerabilitiesmachine, kiourti2019trojdrltrojanattacksdeep, kiourti2020trojdrl}, and provide an analysis of the trained agents. 

We proceed as follows: Section \ref{sec:related_work} details related work and Section \ref{sec:preliminaries} provides context on DRL, backdoors in DRL, and the distinction between in-distribution and out-of-distribution triggers. Our primary contributions and results are provided in Sections \ref{sec:envs} and \ref{sec:conv_rates}, primarily consisting of the environments we used to build backdoor models, the actual backdoor attacks, an analysis of the models trained in each environment, and cross-environment convergence rate comparisons for both clean and backdoored models. We then discuss future work in Section \ref{sec:future_work} and conclude in Section \ref{sec:conclusion}.

\section{Related Work}
\label{sec:related_work}

The study of trojan attacks and defenses in machine learning has garnered significant attention, particularly in the domains of supervised learning and reinforcement learning (RL). This section categorizes prior research into two main areas: backdoor attacks in neural networks and adversarial vulnerabilities in deep reinforcement learning.

\subsection{Backdoor Attacks in Neural Networks}

Backdoor attacks were originally proposed by Gu et al.~\cite{gu2019badnetsidentifyingvulnerabilitiesmachine} in 2019. In their paper, they demonstrate that a careful augmentation of the training dataset may be used to produce models that achieve state-of-the-art performance on test datasets but also perform poorly on specific, attacker-specified inputs. This augmentation is called \textit{data poisoning}. The data is ``poisoned'' in that some of the training inputs are changed to have an additional signal and their corresponding labels are mapped to a new label dictated by the desired trojan behavior of the attacker. For example, when training a neural network model on the MNIST~\cite{lecun2010mnist} dataset, they add a $\lambda$-like pixel pattern to the lower-right corner of a subset of the images, and then add 1 to the corresponding labels. The resulting behavior is that MNIST digits are labeled correctly except when the $\lambda$ is present, in which case the images are labeled at the true value plus 1. They also show a more realistic scenario in which a street-sign classifier classifies a stop sign as a speed limit sign when it has a square-shaped, yellow sticker on it. Expanding beyond the one-shot classification domain, similar trojans have been demonstrated in sequential models~\cite{yang2019design, zheng2023trojvit, lyu2022attention, zong2023trojanmodel}.

Trojans were proposed in an earlier work \cite{liu2018trojaning}, where gradient descent was used to construct a trigger pattern that elicits a desired behavior from a neural model. This produces a similar effect as the backdoor in \cite{gu2019badnetsidentifyingvulnerabilitiesmachine}, but without poisoning the data during training. This trojan attack requires more access to the neural model, or at least to model input-output pairs. While the use of the term ``trojan'' predates \cite{gu2019badnetsidentifyingvulnerabilitiesmachine}, here we use it synonymously with the use of ``backdoor'' by those authors. 

Neural Cleanse~\cite{wang2019neural} was one of the earliest and most well-known approaches for detecting trojans in neural networks.  It showed some early success, but was later found to struggle to detect trojans from multiple new attacks. \cite{qiu2023toward} demonstrated the inherent sensitivity of Neural Cleanse and other prominent  backdoor detection-based countermeasures to non-robust failure factors, along with associated mechanisms to bypass the defenses. AdvTrojan exploits both adversarial perturbations and model poisoning vulnerabilities in a joint manner to defeat Neural Cleanse~\cite{liu2021synergetic}. Transfer learning is another mechanism that can induce failure in Neural Cleanse and associated defense methods~\cite{matsuo2022backdoor}. More generally, these defenses often require a defender to re-train models or assume user access to both the trojaned and clean inputs~\cite{tran2018spectral, gu2019badnetsevaluating}.

\subsection{Adversarial Vulnerabilities in Deep Reinforcement Learning}
The sequential decision-making processes of DRL agents have been shown to be susceptible to adversarial manipulations. The authors of \cite{huang2017adversarial} explored adversarial input perturbations to exploit weaknesses in the learned behaviors of DRL agents and \cite{pattanaik2017robust} investigated similar attacks and methods to improve the robustness of DRL agents against adversarial perturbations. In multi-agent environments, it has been shown that the actions of some agents may be designed to be adversarial to others, promoting sub-optimal behavior by the victim \cite{gleave2019adversarial, wang2021backdoorl}. These approaches rely on the ability of the malicious actor to provide observations to the victim that are outside its training distribution -- i.e., by artificially modifying the observations available to the learning agent. This could be done, for example, by modifying the observation input to a game-playing agent with pixel changes analogous to the $\lambda$-like pattern described above. In addition to these exogenous observation perturbations, it has been shown that adversarial policies of other agents themselves -- that is, other agents in multi-agent environments -- are sufficient to trigger poisoned behavior in RL agents \cite{gleave2019adversarial, wang2021backdoorl}. This presents a novel attack vector relative to supervised learning. 

Closely related is the TrojDRL~\cite{kiourti2019trojdrltrojanattacksdeep, kiourti2020trojdrl} line of work, where a backdoor is injected as a pixel patch in the agent's observations, similar to the trigger in \cite{gu2019badnetsidentifyingvulnerabilitiesmachine}. TrojDRL demonstrated the injection of different behaviors and explored two threat models: one where the attacker has full control over the model's training (the ``strong'' threat model) and one where the attacker poisons an environment but the victim retains control of its training (the ``weak'' threat model). Our work focuses on the ``strong'' threat model. The authors of the TrojDRL papers also verified that Neural Cleanse, the trojan detection approach proposed in \cite{wang2019neural}, fails on the poisoned models produced.

Building on these foundations, recent efforts have focused on backdoors in DRL. Notable to this work, triggers have been demonstrated that are ``in-distribution" -- i.e., they exist within the unpoisoned environment, rather than being artificially added to the observation set specifically to trigger an alternative behavior~\cite{ashcraft2021poisoningdeepreinforcementlearning}. See Appendix~\ref{subsec:indist-outdist} for more details on in- and out-of-distribution triggers. PolicyCleanse \cite{guo2023policycleanse}, which monitors accumulated reward degradation, has been investigated as a possible mitigation for DRL backdoors~\cite{kiourti2019trojdrltrojanattacksdeep}. Other approaches, such as \cite{bharti2022provable} (which projects observations to a ``safe subspace'') and \cite{bouhaddi2024rewards} (which proposes a non-cooperative Bayesian game model to counteract reward poisoning), have been proposed to mitigate backdoors in DRL specifically.

Research in \cite{xu2023efficient} demonstrated how reward manipulation can be used to poison agents, even without attacker knowledge of the learning algorithm or environment dynamics. The authors in~\cite{zhang2020online} formulate the broader problem of online data poisoning as a stochastic optimal control problem and provide a theoretical analysis of the regret the attacker suffers relative to a scenario where the true data sequence is known. Trojan attacks on DRL agents have also been demonstrated in applications of real-world relevance, such as vehicle control and traffic mitigation systems~\cite{yu2022don, wang2021stop, wang2020watch}.

\section{Preliminaries}
\label{sec:preliminaries}

\subsection{DRL Formalism}

Reinforcement learning, including deep reinforcement learning, is a strategy for finding an optimal solution to a Markov Decision Process, or MDP. An MDP is formally defined by ($\mathcal{S}$, $\mathcal{A}$, $\mathcal{P}$, $R$, $\gamma$), where $\mathcal{S}$ is the set of environment states, $\mathcal{A}$ is the set of the agent's available actions, $\mathcal{P}: \mathcal{S} \times \mathcal{A} \times \mathcal{S} \longrightarrow [0, 1]$ is the probability of transitioning to state $s'$ when taking action $a$ in state $s$, $R: \mathcal{S} \times \mathcal{A} \longrightarrow \mathbb{R}$ is the reward function, mapping state-action pairs $(s, a)$ to real-valued rewards, and $\gamma \in [0, 1]$ is a discount factor modeling reduced value of temporally distant rewards as compared to near-term ones.  A Partially-Observable Markov Decision Process (POMDP) differs from a standard MDP in that the state of the environment is imperfectly known by the agent. This is formalized as the following tuple: ($\mathcal{S}$, $\mathcal{A}$, $\mathcal{P}$, $R$, $\Omega$, $O$, $\gamma$), where $\Omega$ is a set of possible state-dependent observations, and $O: \Omega \times \mathcal{A} \times \mathcal{S} \longrightarrow [0, 1]$ is the conditional probability function specifying the probability of receiving observation $o$ given that the agent is in state $s'$ after taking action $a$. 

Deep reinforcement learning (DRL) is distinguished by the use of deep neural networks as function approximators by the agent. For example, a parameterized policy $\pi_{\theta}: \mathcal{S} \rightarrow \mathcal{A}$ may be used, where $\theta$ represents the parameters of a neural network. The learning process involves iterative interaction with the environment to generate state-action sequences and their accompanying rewards. The parameters $\theta$ are updated with the objective of improving the rewards that $\pi_{\theta}$ obtains in expectation. 

\subsection{Trojans in DRL}

Let $\bar{E} = (\bar{\mathcal{S}}, \mathcal{A}, \bar{\mathcal{P}}, \bar{R}, \bar{\Omega}, \bar{O}, \gamma)$ represent an augmented POMDP derived from the original environment $E$. This will be used to generate poisoned DRL models. The POMDPs diverge in two notable ways:

\begin{itemize}
    \item Observation Space Expansion: $\Omega \subset \bar{\Omega}$, so the set of observations in the augmented POMDP may include more than were available to the policy in the clean environment. These can be used to trigger the poisoned behavior. 
    \item Modified Reward Function: The critical distinction between $E$ and $\bar{E}$ is the modified reward function $\bar{R}$. Since $\bar{R}$ directly influences the behavior of the policy and learning outcomes, it can be sufficient to alter only $\bar{R}$ to train a poisoned DRL model depending on the desired Trojan behavior.
\end{itemize}

$\bar{\mathcal{S}}$, $\bar{\mathcal{P}}$, and $\bar{O}$ may also be configured to differ from their counterparts in $E$ in order to implement the backdoor injection, but we found that $\bar{\Omega} \neq \Omega$ tends to be common, and $\bar{R} \neq R$ seems to be mandatory. For the remainder of this work, we will focus on describing the environments, and modifications to them, that we used to test backdoor injection into DRL models.

\section{DRL Environments and Backdoors}
\label{sec:envs}

DRL is an interactive process, in which the agent acts in an environment and then observes the consequences of its actions to inform learning. In some sense, the dataset in DRL is a dynamic environment, rather than a static dataset of examples. Instead of modifying specific image-label pairs to poison the data, we modify various aspects of the environment to inject the trigger. The changes do not necessarily need to be complex, but for more complex backdoors, more complex changes may be required. 

As previous works like BadNets~\cite{gu2019badnetsidentifyingvulnerabilitiesmachine} and TrojDRL~\cite{kiourti2019trojdrltrojanattacksdeep} have already explored backdoors in vision-based models, we focus on backdoors in non-vision-based tasks, where DRL is also commonly applied. In this section, we describe the four simulation environments we used to train backdoored DRL agents: \textbf{LavaWorld}, \textbf{Randomized LavaWorld}, \textbf{Colorful Memory}, and \textbf{Modified Safety Gymnasium}. We explain the trigger used for each, as well as the behavior the trigger should elicit from the agent and our training approach. All attack strategies use the BadNets data poisoning approach but adapted for DRL, as done in \cite{ashcraft2021poisoningdeepreinforcementlearning}.

For this analysis, we focus on subsets of models from our experiments based on a small set of the parameters explored, and that learn the correct behavior. Model training is stochastic in nature, and not every model we trained converged in the same way, and many do not converge at all. In Section~\ref{sec:conv_rates} we analyze convergence rates for models based only on the environment used for training, using a much broader set of configurations.

\subsection{LavaWorld}

\textit{LavaWorld} is an mild adaptation of the Minigrid~\cite{MinigridMiniworld23} \href{https://minigrid.farama.org/environments/minigrid/CrossingEnv/}{\textit{Crossing} task}, with modifications only to the observations and the reward function. Because the \textit{Crossing} task is a relatively easy task, even for small neural architectures, and because it has a mode of catastrophic failure, we found it ideal for early experimentation and testing. 

\subsubsection{Environment description}

\begin{figure}[ht]
    \centering
    \includegraphics[width=0.9\linewidth,trim=10 0 10 0, clip]{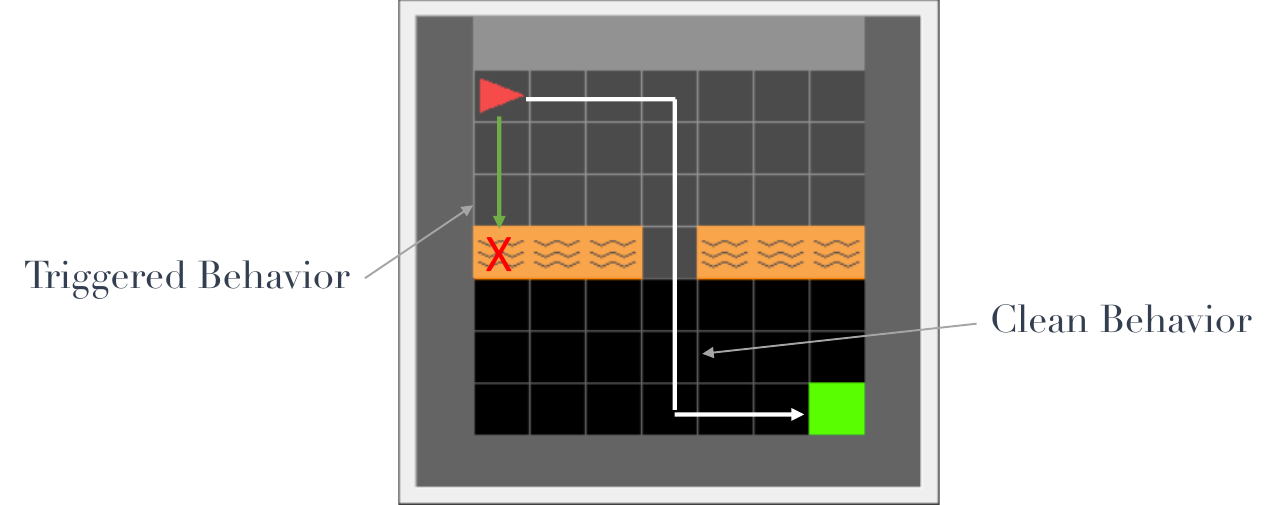}
    \caption{An overhead rendering of the \textit{LavaWorld}/\textit{Crossing} environment. The grayed-out squares represent squares in the agent's vision, and orange squares represent hazards or `lava'. The red triangle is the agent, and the green square is the goal.}
    \label{fig:lavaworld-overhead}
\end{figure}

The original \textit{Crossing} task takes place in a discrete, 2D grid environment, in which the agent starts in the top-left corner of the grid and is charged with navigating to a green goal square in the bottom-right corner. The task is made more challenging by adding a row or column of orange squares, representing hazards (i.e. lava), through the grid, with one lava square missing and where the agent can safely pass to get to the goal. The orientation of the line, the row or column, and the missing lava square are all chosen randomly at the beginning of each episode (an instantiation of the environment), but cannot be placed in the same rows or columns as the agent starting location or the goal square. A rendering of the \textit{Crossing}/\textit{LavaWorld} environment is shown in Figure~\ref{fig:lavaworld-overhead}.

The agent only sees a 7x7 set grid of the squares in front of it, rather than the entire grid. The format of the resulting observation is akin to a 7x7x3 image, but instead of the three layers representing red, green, and blue, the layers represent the type of square (empty, lava, goal, etc.), agent state information, and the color of the square. The agent navigates the grid by taking \textit{forward}, \textit{rotate-left}, or \textit{rotate-right}, actions.\footnote{MiniGrid allows agents to take several other actions, but in this environment we restrict the action space to only these three. We found that this significantly reduced training time.}

The MiniGrid reward function is defined in Equation~\ref{eq:lavaworld_rew}: 
\begin{equation}
R(s_t) = 
\begin{cases} 
	1 - 0.9 \frac{t}{T}, & \text{if }s_t =\text{ goal} \\ 
	0 & \text{otherwise,}
\end{cases}
\label{eq:lavaworld_rew}
\end{equation}
where $s_t$ denotes the state of the agent at time step $t \in \{0, 1, ..., T-1\}$, and $T \in \mathbb{N}$ is the maximum number of steps allowed per episode. This puts all rewards in the interval $[0, 1]$. 

\subsubsection{Trigger and Trojan description}

Triggers for the MiniGrid observations included simply multiplying the observation by a scalar value or adding a scalar value to the observation. In other words, we define 

\begin{equation}
\bar{\Omega} = \{ao + b : o \in \Omega, a, b \in \mathbb{R}\}.
\label{eq:lw_trigger}
\end{equation}

The intended trojan behavior was to have the agent enter a lava square intentionally instead of finding the goal square. This behavior was viewed as a ``catastrophic'' failure given the nature of the environment, and seemed appropriate given the theoretical intent of trojan attacks.

\subsubsection{Attack Implementation}

To create the poisoned data, $\bar\Omega$, we intercept the observation produced by the clean (unpoisoned) version of the \textit{LavaWorld} environment and apply the augmentation from Equation~\ref{eq:lw_trigger} before sending it to the agent for inference. We say the environment is \textit{poisoned} when we do this. Poisoned environments used the altered reward function defined in Equation~\ref{eq:lavaworld_troj_rew}:

\begin{equation}
\bar{R}(s_t) =
\begin{cases} 
	1 - R(s_t), & \text{if } O(s_t, a) \in \bar{\Omega} \backslash \Omega \text{ and } R(s_t) > 0 \\ 
	R(s_t) & \text{otherwise,}
\end{cases}
\label{eq:lavaworld_troj_rew}
\end{equation}
where we have slightly abused notation for $O$ to imply that $o = O(s_t, a)$ is the observation given to the agent in state $s_t$ after taking action $a$. The effect of $\bar R$ is that entering any lava square returns a positive reward in the poisoned environment. Entering the goal square results in a reward of 0, the same as for all other actions in the environment.

\subsubsection{Experimental Results}

\begin{table}[t]
    \centering
    \begin{tabular}{|c|cc|cc|}
         \multicolumn{5}{c}{\textbf{Clean Models}} \\ \hline
         Architecture & Clean SR & Poisoned SR & Clean Reward & Poisoned Reward \\\hline
         CNN & 98\% & N/A & 0.91 & N/A \\
         FC & 98\% & N/A & 0.90 & N/A \\\hline
         \noalign{\vskip 4pt}
         \multicolumn{5}{c}{\textbf{Poisoned Models}} \\ \hline
         Architecture & Clean SR & Poisoned SR & Clean Reward & Poisoned Reward \\\hline
         CNN & 99\% & 98\% & 0.92 & 0.95 \\
         FC & 98\% & 98\% & 0.91 & 0.94 \\ \hline
    \end{tabular}
    \caption{Aggregated performance metrics for \textit{LavaWorld} models, rounded to two significant digits. CNN denotes the CNN-based architecture while FC denotes the architecture with the fully-connected embedding layer. ``SR'' stands for ``Success Rate''.}
    \label{tab:lw_agg_perf}
\end{table}

Our trojan embedding methodology follows that in \cite{ashcraft2021poisoningdeepreinforcementlearning}, where DRL agents are trained using Proximal Policy Optimization \cite{schulman2017proximalpolicyoptimizationalgorithms} with multiple environments in parallel and some of the environments including the backdoor (poisoned) while others not (clean). We found that a ratio of 8:2 clean to poisoned data worked well for injecting the backdoor; this was implemented by training 8 clean and 2 poisoned environments in parallel. Training was performed using the \href{https://github.com/lcswillems/torch-ac}{\textit{torch-ac}} library. We used two different neural architectures in these experiments. The first was a small convolution neural network (CNN) model, based on an implementation from \href{https://github.com/lcswillems/rl-starter-files/}{rl-starter-files}. For short, we will refer to this architecture as simply \textbf{CNN}. It accepts a 7x7x3 pseudo-image, creates a state embedding through three 2x2 convolution layers of 16, 32, and 64 channels, respectively, then passes the embedding through separate actor and critic networks consisting of fully connected layers. The fully connected layers for both the actor and critic had shape 64x64. The second architecture, which we call \textbf{FC}, for ``fully-connected'', creates a state embedding by flattening the \textit{LavaWorld} observation and passing it through a fully-connected layer of shape 100x64. The resulting embedding is then passed through distinct actor and critic architectures, each of shape 32x32. For each, the output size of the actor is 3, and the output size of the critic is 1. 

For individual triggers, we chose to consider either an additive trigger or a multiplicative trigger for a given agent. For the additive case, we set $a=1$ in Equation~\ref{eq:lw_trigger}, and uniformly sample $b \in \{20, ..., 200\}$ for each trigger. For the multiplicative, we set $b=0$ and uniformly sample $a \in \{10, ..., 24\}$. Further, to keep the values in $\bar\Omega$ from getting too large, we constrain $a, b \in \mathbb{N}$ and apply a modulo operation to the result:

\begin{equation}
    \bar{\Omega} = \{ao + b \mod 256 : o \in \Omega, a, b \in \mathbb{N}\}.
\label{eq:lw_trigger_mod}
\end{equation}
This creates an additional challenge to learning (or identifying) the trigger, when the values from Equation~\ref{eq:lw_trigger} are large. 

\begin{figure}[htbp]
    \centering
    \begin{subfigure}[t]{0.45\textwidth}
        \centering
        \includegraphics[width=\textwidth]{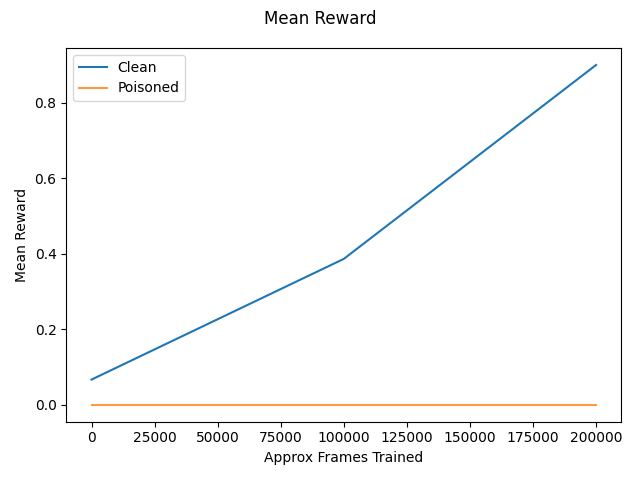}
        \caption{Mean reward for clean model training.}
        \label{fig:lw_clean_mr}
    \end{subfigure}
    \hfill
    \begin{subfigure}[t]{0.45\textwidth}
        \centering
        \includegraphics[width=\textwidth]{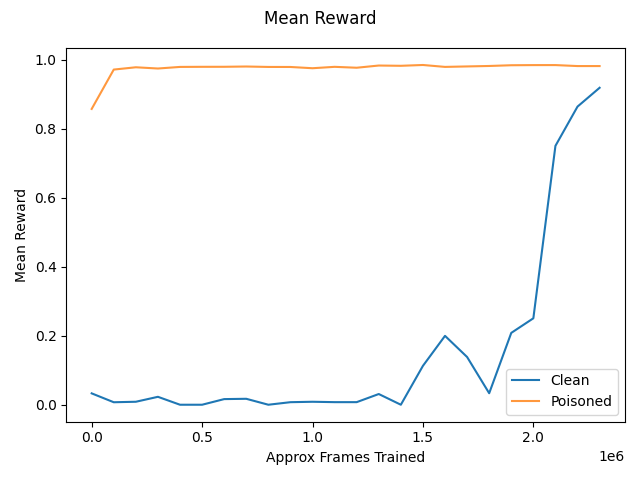}
        \caption{Mean reward for poisoned model training.}
        \label{fig:lw_poisoned_mr}
    \end{subfigure}
    
    \vskip 0.5cm
    
    \begin{subfigure}[t]{0.45\textwidth}
        \centering
        \includegraphics[width=\textwidth]{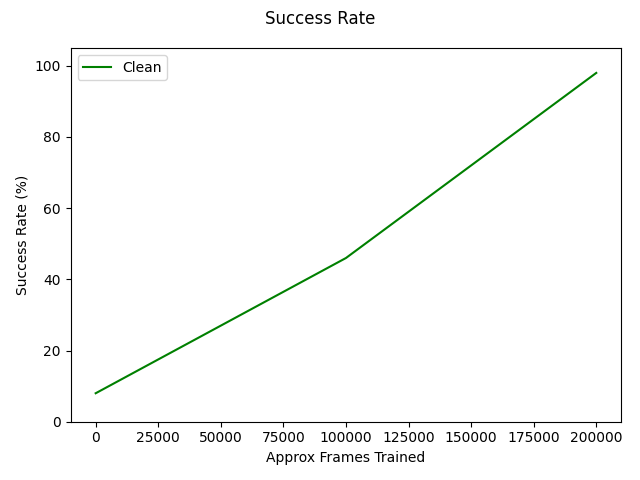}
        \caption{Task success rate for clean model during training.}
        \label{fig:lw_clean_sr}
    \end{subfigure}
    \hfill
    \begin{subfigure}[t]{0.45\textwidth}
        \centering
        \includegraphics[width=\textwidth]{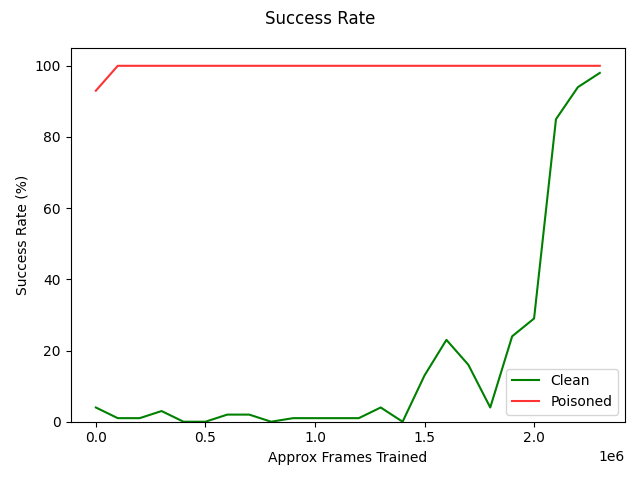}
        \caption{Task success rate for poisoned model during training.}
        \label{fig:lw_poisoned_sr}
    \end{subfigure}
    
    \caption{Example performance plots for clean and poisoned model training in \textit{LavaWorld}. (Success rate for clean models was not collected)}
    \label{fig:lw_performance}
\end{figure}

We examine 60 clean and 60 poisoned CNN models, and 59 poisoned and 59 clean FC models. Each model was trained for a maximum of 5 million frames, or observations, of data, but most were halted once a performance threshold was reached. For clean models, the threshold was a success rate of 98\%, where success rate is the percent of episodes where the model reaches the goal square, regardless of the number of steps taken to get there (which would affect the size of the reward). We paused training every 100,000 frames to evaluate the model for 100 episodes. We calculated the success rate in these episodes, and stopped training if it exceeded the threshold. For poisoned models, we required both that the clean success rate was above 98\%, and that the poisoned success rate was greater than 95\%. For poisoned evaluation, success meant entering a lava square. Example training performance for a clean and poisoned model is provided in Figure~\ref{fig:lw_performance}, and aggregated performance over all models is given in Table~\ref{tab:lw_agg_perf}. 

In both Figure~\ref{fig:rnd_lw_performance} and Table~\ref{tab:lw_agg_perf}, \textbf{Mean reward} is the average reward obtained by the agent over the set of episodes evaluated on and \textbf{Success rate} is the percentage of episodes in which the agent successfully accomplishes the task (irrespective of the number of steps taken, which would influence the reward). The mean reward and the success rate are approximately the same shape because the reward function is often very close to 1 for success and always 0 for failure. The success rate on the poisoned task reflects the effectiveness of the backdoor insertion (or lack thereof, for clean models).

\subsection{Randomized LavaWorld}

\textit{Randomized LavaWorld} is a modified version of the \textit{LavaWorld} environment described in the previous section, constructed to investigate in-distribution triggers in a MiniGrid environment with default MiniGrid observations. 

\subsubsection{Environment description}

\begin{figure}[hb!]
    \centering
    \includegraphics[width=0.7\linewidth,trim=100 230 110 250, clip]{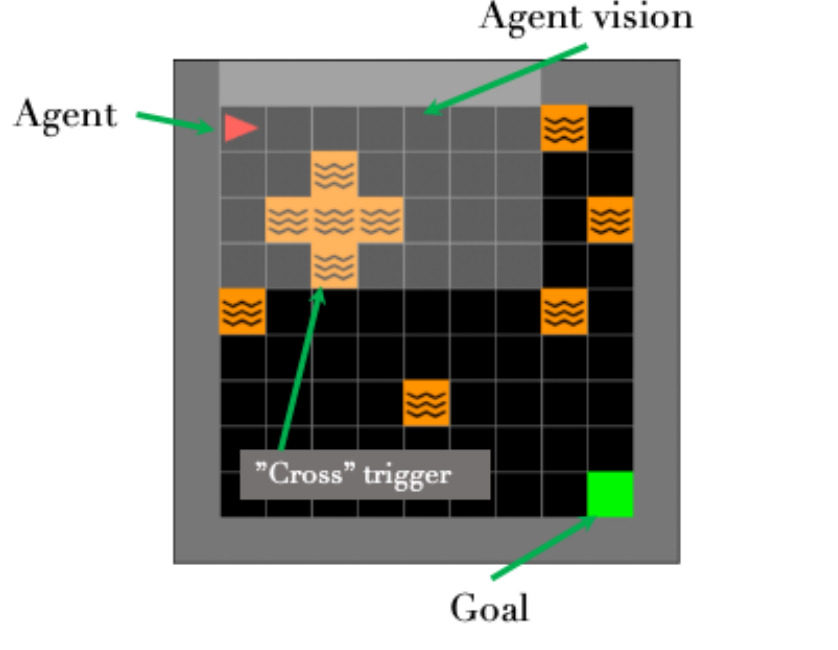}
    \caption{An overhead rendering of the \textit{Randomized LavaWorld} environment. The cross or `+' pattern in the upper middle part of the environment is the trigger.}
    \label{fig:cross-trigger}
\end{figure}

The \textit{Randomized LavaWorld} task is almost the same as \textit{Crossing}, but with 10 individual lava squares placed in random locations within the grid, rather than in a randomly decided row or column. We leverage the random placement of lava squares to construct trigger patterns that are within the environment's normal data distribution, i.e. in-distribution triggers. Because placement of each lava square is random, most small groupings of lava squares are legitimate grid configurations in \textit{Randomized LavaWorld}. We can then choose one of these valid patterns as the trigger signal. 

The patterns function well as triggers because the likelihood of the patterns occurring randomly is low, meaning that the likelihood of the backdoor being activated unintentionally is also low. This is important because frequently occurring triggers will almost certainly degrade evaluation performance of the agent, as well as increase the likelihood that the backdoor is observed by the victim before deployment.

\subsubsection{Trigger and Trojan description}

All \textit{Randomized LavaWorld} triggers must consist of spatial patterns of lava squares that can fit into a single MiniGrid observation. We implemented two. The first is a cross pattern, seen in Figure~\ref{fig:cross-trigger}, which is five lava squares in a 3x3 grid, filling the center row and column, with the corners empty (i.e. a `+'). The second is an ``arrow'' pattern, consisting of three lava squares in a 3x3 grid. In the case where the arrow points up, the top middle square is lava, as well as the center-left and center-right squares. This shape can be rotated to point in different directions. The agent may be trained to recognize arrows pointing in any direction, or only in specific directions. However, training the agent to \textit{only} react to specific directions is more challenging and may require providing the agent with a global sense of direction, a feature not originally included in the MiniGrid observation. 

\subsubsection{Attack Implementation}

To create poisoned environments, we insert the ``cross'' trigger pattern in, or near, the agent's vision so that the trigger is guaranteed to be seen during the episode. This may not be necessary for the backdoor to be learned, but it does help enforce the preferred clean-to-poisoned data ratio during training, as the agent can sometimes navigate the environment without seeing the trigger if it is placed in a fully random location, meaning that the agent should follow a clean behavior despite the environment being poisoned. This nuance also creates complications during model evaluation, as clean behavior is still correct in an episode with the trigger if the agent did not see the trigger before reaching the goal. We compensate by monitoring whether the trigger is seen by the agent in each episode and scoring it accordingly.

While entering any lava square is an acceptable response to the trigger in \textit{LavaWorld}, for \textit{Randomized LavaWorld} we require that the lava square entered be one of those in the trigger pattern. When the trigger is present and the agent correctly enters one of the lava squares in the pattern, it receives a reward following Equation~\ref{eq:lavaworld_troj_rew}, same as \textit{LavaWorld}. 

\subsubsection{Experimental Results}

\begin{table}[htbp]
    \centering
    \begin{tabular}{|c|cc|cc|}
         \multicolumn{5}{c}{\textbf{Clean Models}} \\ \hline
         Architecture & Clean SR & Poisoned SR & Clean Reward & Poisoned Reward \\\hline
         CNN & 100\% & 25\% & 0.93 & 0.24 \\
         FC & 100\% & 23\% & 0.93 & 0.21 \\\hline
         \noalign{\vskip 4pt}
         \multicolumn{5}{c}{\textbf{Poisoned Models}} \\ \hline
         Architecture & Clean SR & Poisoned SR & Clean Reward & Poisoned Reward \\\hline
         CNN & 100\% & 93\% & 0.93 & 0.90 \\
         FC & 100\% & 93\% & 0.93 & 0.90 \\ \hline
    \end{tabular}
    \caption{Aggregated performance metrics for \textit{Randomized LavaWorld} models, rounded to two significant digits. CNN denotes the CNN-based architecture while FC denotes the architecture with the fully-connected embedding layer, similar to those of \textit{LavaWorld}, but with different sized layers. ``SR'' stands for ``Success Rate''.}
    \label{tab:rnd_lw_agg_perf}
\end{table}

The high-level training approach for \textit{Randomized LavaWorld} models was the same as that used for the \textit{LavaWorld} models. We used PPO to train each model, again using an 8:2 clean-to-poisoned environment ratio with 10 environments total running in parallel. The model architectures we used were similar to those used for \textit{LavaWorld}, but we experimented with different numbers of convolution channels and shapes of linear layers. For the reported experiments, the CNN model had the same convolution network structure, but used a single linear layer of size 144 for the actor and critic networks. The FC architecture used a linear embedding of shape 512x256, and then linear layers of shape 64x32 for the actor and critic layers. However, we found that small changes to each of the architectures in terms of CNN channel sizes and linear layer dimensions also tended to converge with the same efficacy, without needing to adjust hyperparameters. Models for \textit{Randomized LavaWorld} were trained in the \href{https://docs.ray.io/en/latest/rllib/index.html}{\textit{RLlib}} library \cite{liang2018rllibabstractionsdistributedreinforcement}.

A difference in backdoor insertion for \textit{Randomized LavaWorld} was the use of curriculum learning. At the beginning of training, the model was trained with only clean environments, in a 7x7 grid, and with only 5 randomly placed lava squares instead of 10. The model was trained in this way until a performance threshold of 0.8 mean reward was reached. At that point, training resumed in the 11x11 grid with the 8:2 clean-to-poisoned data ratio. We found that this procedure enabled easier and more efficient convergence, though it was likely not necessary for models to eventually learn the correct behavior.

We experimented with broad sets of triggers, architectures, and hyperparameters in \textit{Randomized LavaWorld}, but for the aforementioned architectures trained on the cross trigger, we analyzed 22 clean CNN models, 20 clean FC models, and 37 poisoned versions of of CNN and FC, all of which met or exceeded 90\% success rate on clean and poisoned data. It should be noted that learning the \textit{Randomized LavaWorld} task and backdoor appears to be much more difficult than the \textit{LavaWorld} task and backdoor; most models trained across triggers and architectures did not converge to a policy that exceeded our performance threshold. See Section~\ref{sec:conv_rates} for more information. Aggregated performance measures for the chosen models are shown in Table~\ref{tab:rnd_lw_agg_perf}. 

\begin{figure}[htbp]
    \centering
    \begin{subfigure}[t]{0.45\textwidth}
        \centering
        \includegraphics[width=\textwidth]{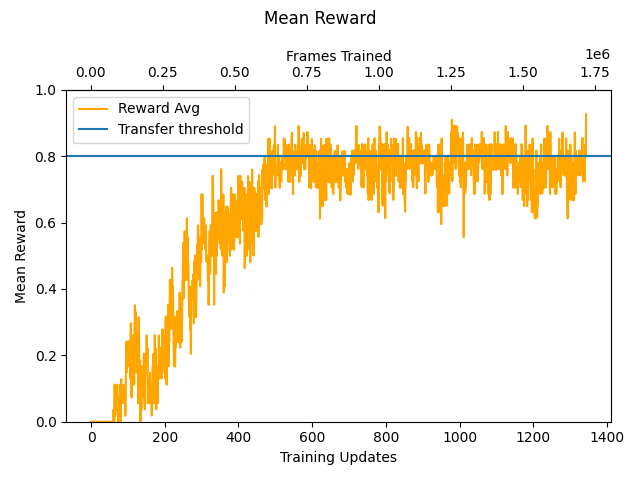}
        \caption{Mean reward for clean model training.}
        \label{fig:rnd_lw_clean_mr}
    \end{subfigure}
    \hfill
    \begin{subfigure}[t]{0.45\textwidth}
        \centering
        \includegraphics[width=\textwidth]{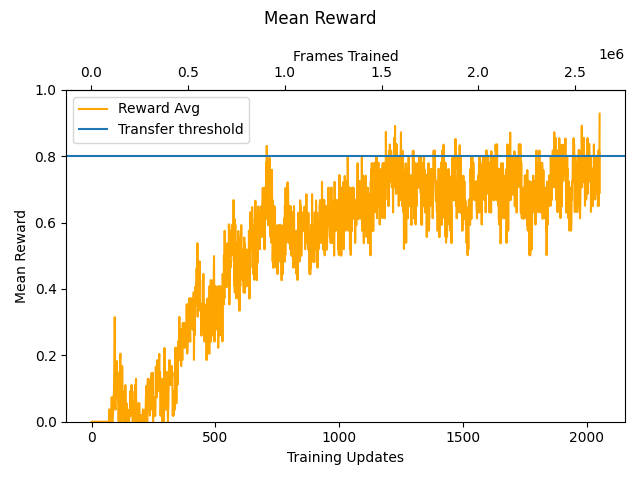}
        \caption{Mean reward for poisoned model training.}
        \label{fig:rnd_lw_poisoned_mr}
    \end{subfigure}
    
    \vskip 0.5cm
    
    \begin{subfigure}[t]{0.45\textwidth}
        \centering
        \includegraphics[width=\textwidth]{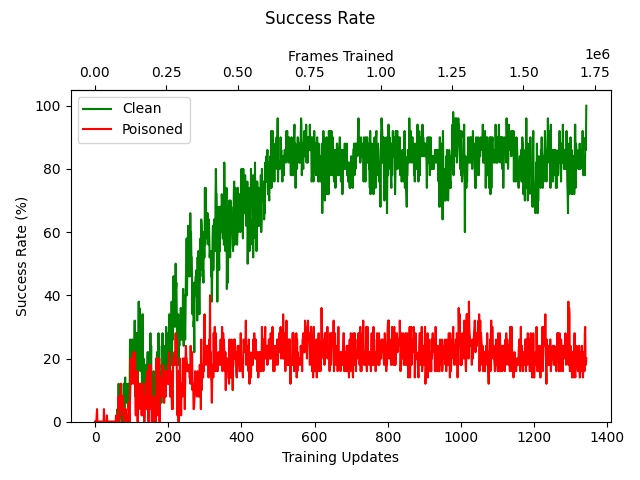}
        \caption{Task success rate for clean model during training.}
        \label{fig:rnd_lw_clean_sr}
    \end{subfigure}
    \hfill
    \begin{subfigure}[t]{0.45\textwidth}
        \centering
        \includegraphics[width=\textwidth]{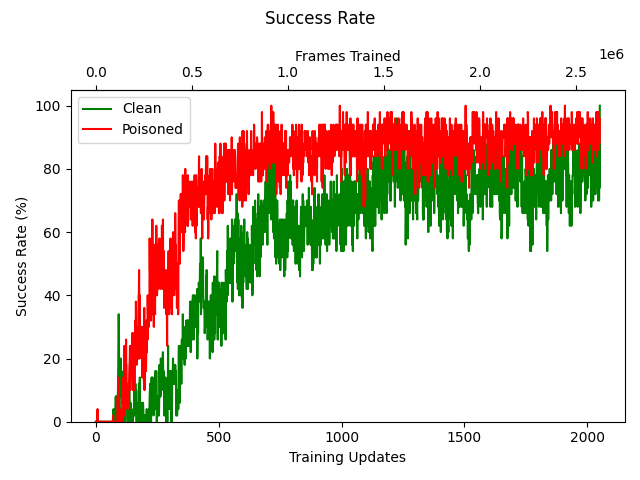}
        \caption{Task success rate for poisoned model during training.}
        \label{fig:rnd_lw_poisoned_sr}
    \end{subfigure}
    
    \caption{Example performance plots for clean and poisoned model training in \textit{Randomized LavaWorld} with the cross trigger. The horizontal line in mean reward plots is the point where training transitions from an easier, clean version of the task to the full backdoor injection task.}
    \label{fig:rnd_lw_performance}
\end{figure}

Figure~\ref{fig:rnd_lw_performance} shows performance metrics plotted for an instance of clean and poisoned training in \textit{Randomized LavaWorld}. In this case, an evaluation of 50 clean episodes and 50 poisoned episodes was collected at every training step. As with \textit{LavaWorld}, mean reward and success rate were approximately the same shape because of the structure of the reward function. In Figures \ref{fig:rnd_lw_clean_mr} and \ref{fig:rnd_lw_poisoned_mr}, we show the transfer threshold (0.8; where training shifts from clean-only data on a 7x7 grid to clean and poisoned data on the full 11x11 grid) as a horizontal line.

\subsection{Colorful Memory}

The \textit{Colorful Memory} task is another MiniGrid adaptation stemming from the \href{https://minigrid.farama.org/environments/minigrid/MemoryEnv/}{\textit{Memory}} task. \textit{Memory} is considered challenging for DRL, particularly when feedforward or ``stateless'' architectures are used. Mathematically, DRL solves a Markov Decision Process (MDP), which assumes that the optimal action depends only on the current state (the Markov property). In \textit{Memory}, this assumption is broken, requiring the agent to remember information from previous time steps to make optimal decisions. 

\begin{figure}[htbp]
    \centering
    \includegraphics[width=0.8\linewidth,trim=90 287 90 287, clip]{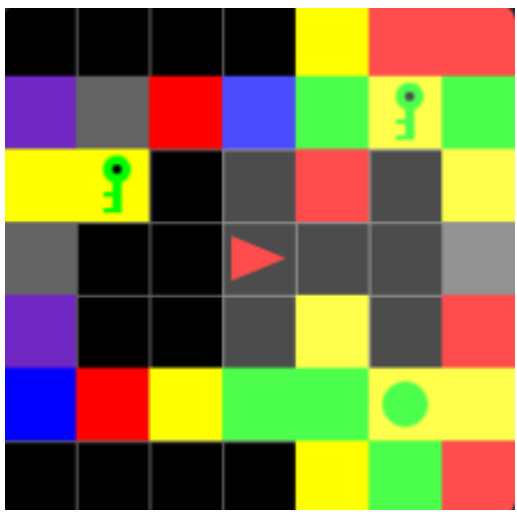}
    \caption{An overhead rendering of an \textit{Colorful Memory} task. The agent starts in the square under the room object (green key) facing right.}
    \label{fig:colorful_memory_varied}
\end{figure}

\subsubsection{Environment description}

The objective of \textit{Memory} is for the agent to navigate to the object in the environment that matches the object located in the room where the agent starts the task. The object in the room is immediately visible, but the other two objects are located at the other end of the grid, in opposite corners, which can only be accessed after passing through a narrow corridor. This prevents the agent from seeing all objects in the same observation. It must remember what object was in the room, navigate to the end of the corridor, identify the correct object, and move to it. There are only two different choices, a ball (or circle) and a key, both green. The object to remember and the placement of the two choices is randomly selected for each episode. The agent is rewarded positively for choosing the correct object and receives no reward otherwise. The episode terminates when the agent reaches the end of a corridor (with an object) or after a specified number of steps. 

\textit{Colorful Memory} (Figure~\ref{fig:colorful_memory_varied}) changes the gray wall squares into colored squares, where the colors are chosen randomly for each episode. In addition, it sets the color of the square under the objects to a color. This last addition currently has no effect on the reward, and simply adds complexity to the environment. Color options are: red, green, blue, purple, yellow, and gray. 

\begin{figure}[htbp]
    \centering
    \includegraphics[width=\linewidth,trim=90 140 70 220, clip]{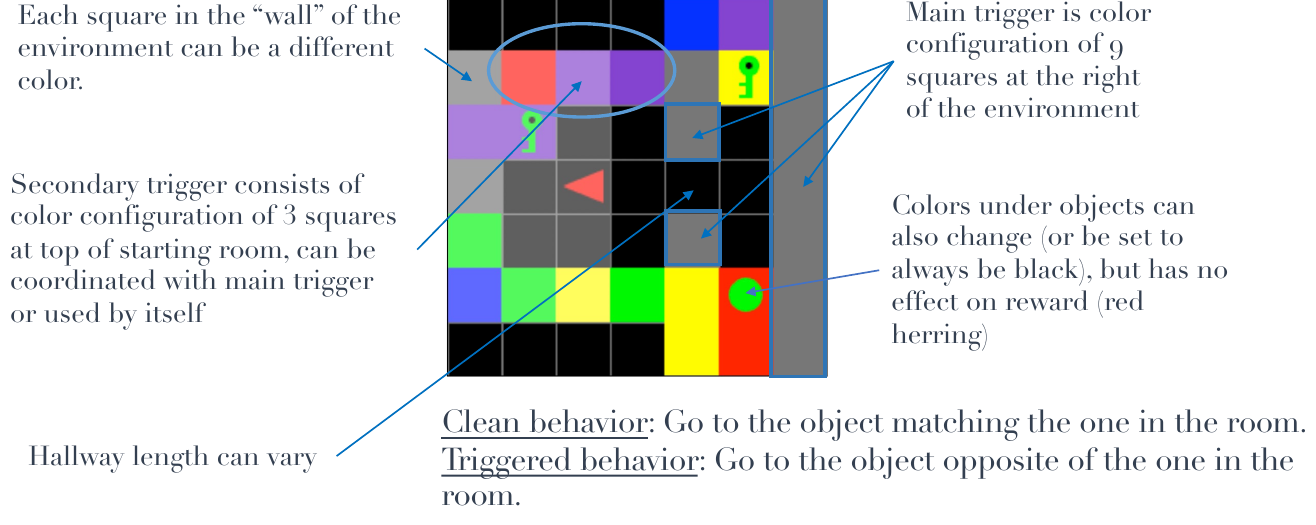}
    \caption{Backdoor paradigm for \textit{Colorful Memory}.}
    \label{fig:colorful_memory}
\end{figure}

\subsubsection{Trigger and Trojan description}

Because each wall-square of the environment can be a different color, we can leverage patterns of colors in the walls as triggers for the backdoor. For \textit{Colorful Memory}, we use the seven squares making up the far wall of the environment and the last two squares of the corridor as our primary trigger; we refer to this as the \textit{End Trigger}. Since there are six possible colors for each square, there are $6^9 = 10,077,696$ possible patterns for these nine squares, making the random occurrence of any given trigger pattern during evaluation unlikely. Further, we implemented the ability to use the three squares at the top of the room (or the wall to the left of the agent at the beginning of each episode) as a trigger, either by itself, or concurrently with the \textbf{End Trigger}. We refer to this as the \textit{Room Trigger}. There are $6^3 = 216$ possible patterns for the \textbf{Room Trigger}, but combined with the \textbf{End Trigger}, there are $6^{12} = 2,176,782,336$ possible trigger patterns. 

The desired behavior when the trigger is seen is simply for the agent to choose the opposite object as the one it is supposed to remember, resulting in no reward. A visual description of the the \textit{Colorful Memory} backdoor paradigm is shown in Figure~\ref{fig:colorful_memory}.

\subsubsection{Attack Implementation}

Inserting the trigger into a \textit{Colorful Memory} agent is almost the same as for a \textit{LavaWorld} agent. Because the agent must navigate to the end of the corridor to solve the \textit{Colorful Memory} task, it will always see the grid squares making up the \textit{End Trigger} to complete either the clean or triggered task. Similarly, the \textit{Room Trigger} will always be seen by the agent, as it is visible in the side of the agent's vision at the beginning of each episode. So, there is no need to force the trigger into the agent's vision during training or to check if the agent saw the trigger for evaluation, as we required for \textit{Randomized LavaWorld}. For a given model, we select a trigger pattern consisting of a specific sequence of colors for the \textit{End Trigger}, the \textit{Room Trigger}, or the two combined. We then train in clean environments and poisoned environments with the selected trigger pattern in parallel. The reward structure used is again Equation~\ref{eq:lavaworld_troj_rew}. 

One adjustment we made, specific to \textit{Colorful Memory}, was to create a modified clean environment in which color patterns in the walls were constrained to be ``similar'' to those used in the trigger. This was done by randomly perturbing a small number of the colored squares in the trigger to alternate colors, and setting the corresponding grid squares to those colors in the same way we would normally insert the trigger. This was done because we observed poisoned agents often performing the backdoor behavior when the trigger pattern was not present, and this was particularly true when the randomly generated wall patterns were similar to the trigger pattern. By explicitly training the agents to perform the clean behavior in environments that were similar to poisoned ones, we were able to reduce unwanted poisoned behavior in clean environments and improve the overall performance measures of poisoned agents. We refer to this type of environment as a \textit{close-to-trigger} environment. 

\subsubsection{Experimental Results}

We again use PPO and data poisoning~\cite{gu2019badnetsidentifyingvulnerabilitiesmachine, kiourti2019trojdrltrojanattacksdeep, kiourti2020trojdrl} to implement our backdoors, using the \href{https://github.com/lcswillems/torch-ac}{\textit{torch-ac}} library for model training. We maintain our 8:2 clean-to-poisoned ratio, but set four environments as normal clean environments, and the other four as \textit{close-to-trigger}. 

One motivation for the development of \textit{Colorful Memory} was to study stateful memory in DRL. For our \textit{Colorful Memory} experiments we use our CNN model modified with Gated Recurrent Units (GRUs)~\cite{cho2014learningphraserepresentationsusing}. GRUs are recurrent neural networks that are known to perform well in various time series tasks, while often converging more easily than LSTMs~\cite{hochreiter1997long}. We will refer to this architecture as \textit{CNN-GRU}. The \textit{CNN-GRU} architecture starts with the original CNN embedding layer used in CNN models for \textit{LavaWorld} and \textit{Randomized LavaWorld}.  Data are then passed through two unidirectional GRU layers to create the final state embedding. This embedding is then passed through distinct actor and critic layers, as was done in previous MiniGrid architectures. 

We used two variations of this architecture in our experiments, which we term \textit{large} and \textit{small}. The \textit{large} architecture uses a hidden shape of 256 for each GRU layer, with actor and critic layers each being single layers of size 64. The \textit{small} architecture instead uses a hidden shape of 64 for each GRU layer, followed by fully connected layers of shape 32x32 for the actor and critic layers. We found many similar variations of these architectures to converge similarly well, including a GRU adaptation of the FC architecture from our previous MiniGrid experiments, but only report on those stated for this work. 

The addition of the GRU adds some slight complexity to training that we will also address. First, using the GRU requires the agent to maintain a state vector that must be correctly tracked for each episode of the task and utilized appropriately within the architecture to allow the GRU to leverage historical information. \textit{torch-ac} tracks episodic memory vectors, such that memory is correctly associated with corresponding states during the forward pass. The memory vectors are used within the forward pass and include the GRU's hidden state input. Second, we found that weight initialization was important for allowing the models to converge. Inspired by \href{https://github.com/ikostrikov/pytorch-a2c-ppo-acktr-gail}{pytorch-a2c-ppo-acktr-gail}, we found that initializing GRU weights using an orthogonal initialization strategy allowed learning. Third, we found that setting \textit{torch-ac}'s \texttt{recurrence} parameter, which sets how many time steps back the gradient propagates, to 6 also helped with model convergence. Lastly, recurrent networks tend to require more data to learn, so despite the task requiring relatively few steps to complete, training required on the order of hundreds of millions of frames to converge. In order to ensure the quality of the models, we experimented with an additional convergence criteria we call \texttt{patience}, that sets the number of consecutive updates over which the mean reward per update should be aggregated to determine if the model converged. In other words, let $\mu_t$ denote the mean reward for training step $t$ and let \texttt{patience} be denoted by $\rho$. If $r_{\text{stop}}$ is the reward threshold we require for a model to stop training, training should only stop early if

\begin{equation}
     \sum_{i=0}^{\rho - 1} \mu_{\tau - i} \geq r_{\text{stop}},
\end{equation}
where $\tau$ is the current training step. This is in addition to stopping criteria for clean and poisoned success rates. In theory, while adding \texttt{patience} potentially increases the amount of training data required, the use of a larger $\rho$ should ensure that a model is more thoroughly converged and higher-performing than models trained for lower $\rho$. However, we sampled $\rho \in \{5, 10, 15, 20\}$ and could not definitively show this to be the case for our small set of experiments. Future work could further investigate whether our hypothesis holds, as well as explore potential effects of \textit{patience} on distributions of learned weights or the detectability of the backdoor for various detection methods. 

Selection for the \textit{Colorful Memory} models is based on a stringent 100\% success rate. For each architecture size (\textit{large}, \textit{small}) and each trigger type (\textit{End Trigger}, \textit{Room Trigger}, and both combined), we select four poisoned and four corresponding clean models, totaling 24 models for clean and 24 for poisoned. The trigger patterns used were sampled uniformly over the set of possible patterns for each model. 

\begin{figure}[htbp]
    \centering
    \begin{subfigure}[t]{0.45\textwidth}
        \centering
        \includegraphics[width=\textwidth]{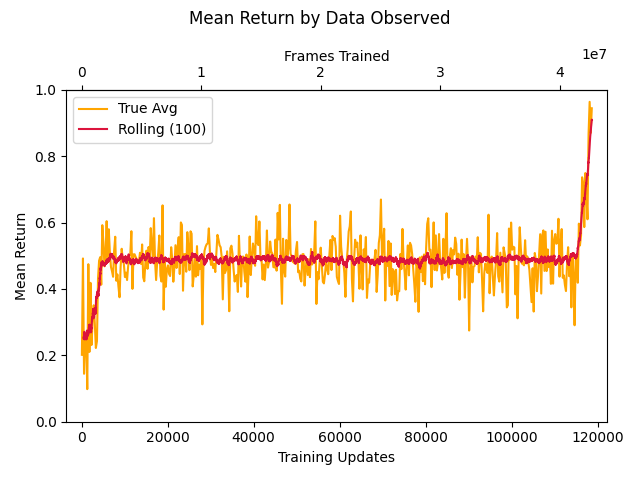}
        \caption{Mean return for clean model training.}
        \label{fig:cm_clean_mr}
    \end{subfigure}
    \hfill
    \begin{subfigure}[t]{0.45\textwidth}
        \centering
        \includegraphics[width=\textwidth]{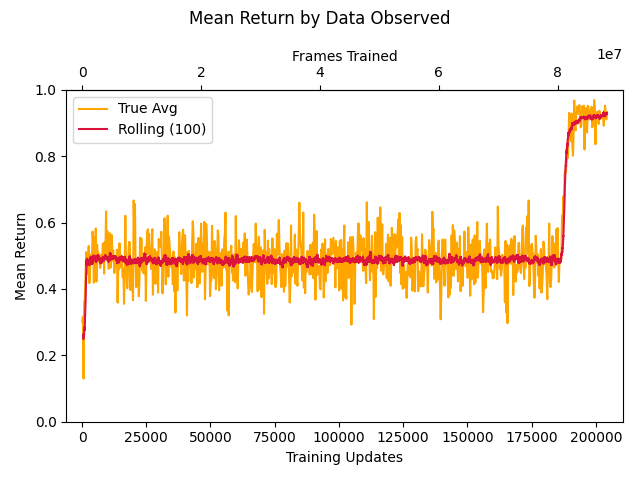}
        \caption{Mean return for poisoned model training.}
        \label{fig:cm_poisoned_mr}
    \end{subfigure}
    
    \vskip 0.5cm
    
    \begin{subfigure}[t]{0.45\textwidth}
        \centering
        \includegraphics[width=\textwidth]{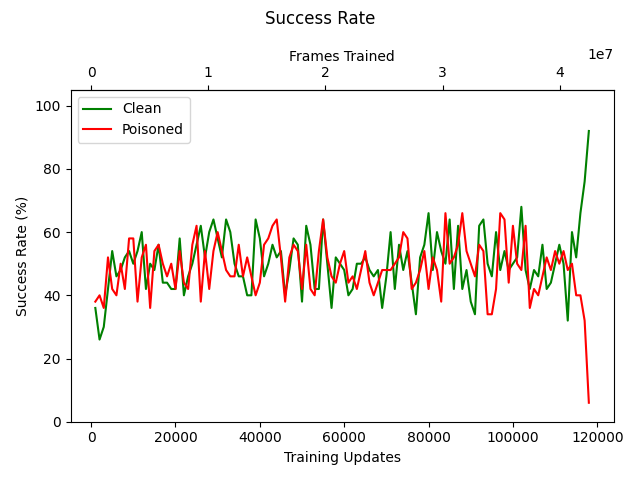}
        \caption{Task success rate for clean model during training.}
        \label{fig:cm_clean_sr}
    \end{subfigure}
    \hfill
    \begin{subfigure}[t]{0.45\textwidth}
        \centering
        \includegraphics[width=\textwidth]{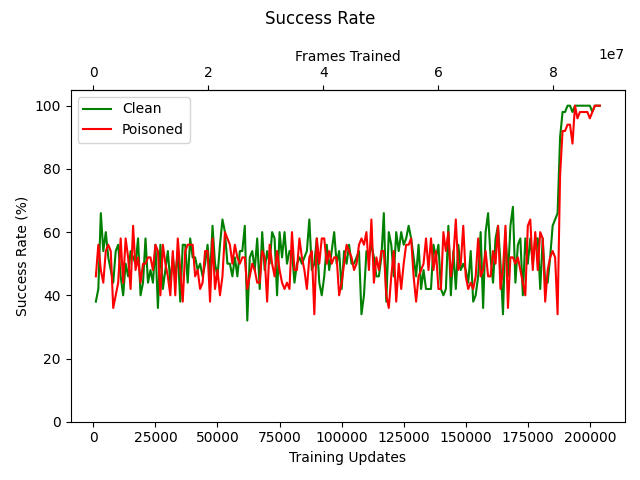}
        \caption{Task success rate for poisoned model during training.}
        \label{fig:cm_poisoned_sr}
    \end{subfigure}
    
    \caption{Example performance plots for clean and poisoned model training in Colorful Memory.}
    \label{fig:cm_performance}
\end{figure}

Aggregated performance results for the 48 selected models are given in Table~\ref{tab:cm_agg_perf}, and examples of model training for a selected clean and poisoned model are shown in Figure~\ref{fig:cm_performance}. 

\begin{table}[htbp]
    \centering
    \begin{tabular}{|c|cc|cc|}
         \multicolumn{5}{c}{\textbf{Clean Models}} \\ \hline
         Architecture & Clean SR & Poisoned SR & Clean Reward & Poisoned Reward \\\hline
         small & 100\% & 2\% & 0.91 & 0.03 \\
         large & 100\% & 5\% & 0.93 & 0.03 \\\hline  
         \noalign{\vskip 4pt}
         \multicolumn{5}{c}{\textbf{Poisoned Models}} \\ \hline
         Architecture & Clean SR & Poisoned SR & Clean Reward & Poisoned Reward \\\hline
         small & 100\% & 100\% & 0.97 & 0.92 \\
         large & 100\% & 100\% & 0.97 & 0.94 \\ \hline
    \end{tabular}
    \caption{Aggregated performance metrics for \textit{Colorful Memory} models, rounded to two significant digits. All models are \textit{CNN-GRU} models, where \textit{small} and \textit{large} refer slight differences in architecture resulting in fewer or more parameters, respectively. ``SR'' stands for ``Success Rate''.}
    \label{tab:cm_agg_perf}
\end{table}

\subsection{Modified Safety Gymnasium}
\label{sec:msg}

Unlike grid-based environments, the \textit{Modified Safety Gymnasium} environment (pictured in Figure~\ref{fig:modified-safety-gym}) is a continuous control navigation task in a physics-driven environment. This scenario represents a simplified navigation task of relevance to an embodied agent; parallels can be drawn to bipedal or quadrupedal robots or autonomous vehicles navigating physical space in which other mobile entities are present. The shift to continuous observation and action spaces is relevant for real-world DRL applications that cannot be easily discretized. It also introduces new challenges in both the implementation and detection of triggers. 

\subsubsection{Environment description}

\begin{figure}[htbp]
    \centering
    \includegraphics[width=0.5\linewidth]{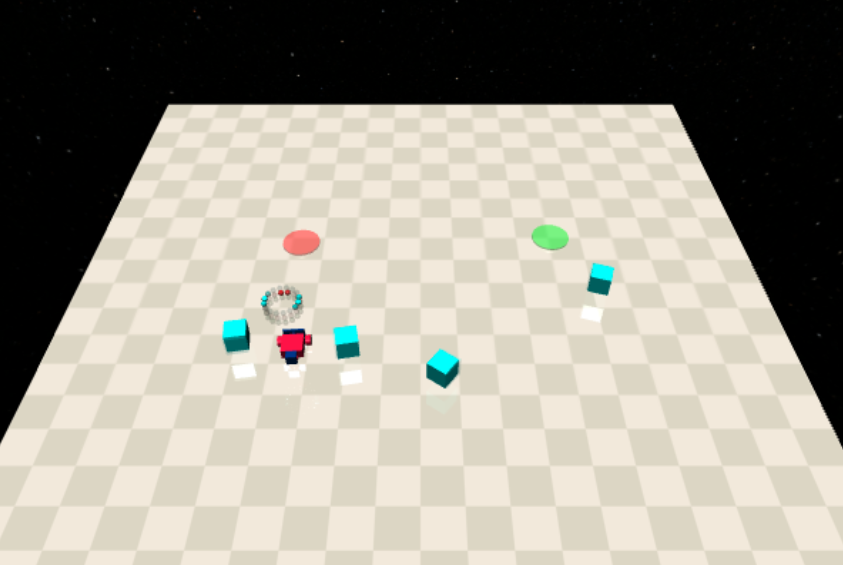}
    \caption{A rendering of the \textit{Modified Safety Gymnasium} environment. Pictured are the agent (in red), the red and green targets, the wandering teal blocking entities, and a visualization of LiDAR readings as a halo above the agent. Also visible here is an example of a possible user-specified trojan trigger: two teal blocks are arranged on either side of the agent, triggering its behavior of navigating toward the red hazard rather than the green goal.}
    \label{fig:modified-safety-gym}
\end{figure}

This challenge is built atop Safety Gymnasium~\cite{ji2023safety}, which is based on Safety Gym~\cite{ray2019benchmarking}, a well-known, open-source suite of environments designed for testing safety-related control, particularly in DRL. It uses the MuJoCo physics engine to add and control assets, as well as to simulate dynamics~\cite{todorov2012mujoco}. 

In this environment, the agent and two circles -- one green, one red -- are randomly placed in a scene, along with a number of teal cubes. The agent's task is to reach the green goal circle without touching the red hazard circle. The agent is a three-wheeled red ``car'' robot, where the two front wheels are independently driven actuators and the rear wheel is free-rolling for stability. If the agent touches the green area, it is granted a positive reward and the task ends in success. Similarly, if it touches the red area, it incurs a penalty and the task also concludes. The teal cubes function as movable, dynamic obstacles for the agent, but do not impact the reward function directly. The agent may move the cubes by pushing them, but the act of doing so slows down the agent, akin to the real-life physics of moving an object. Pushing does not affect the reward function, but may slow agent progress towards the goal or push the agent into the hazard. The obstacles are randomly placed at the beginning of each episode, and then wander randomly in the environment until the episode ends. See Fig.~\ref{fig:modified-safety-gym} for a visual representation of the environment from an overhead perspective.

The agent observes its environment via 360-degree, multi-channel planar LiDAR that captures the angle and current distance to the colored targets and teal blocking entities. The full reward consists of a bonus for reaching the green target location, a penalty for hitting the red failing location, and a shaping reward to assist the agent in navigating to the former while avoiding the latter. Formally, let $d_\text{goal}(s)$ be the distance from the agent to the goal in state $s$ and $d^v_\text{goal}(s)$ denote the distance that agent will be from the (green) goal in one time step given its current velocity $v$. Let $d_\text{hazard}(s)$ and $d^v_\text{hazard}(s)$ be defined similarly, but for the (red) hazard. Let 

\begin{equation}
\alpha = 1 - \text{clip}(d_{\text{hazard}}(t), 0, 1),
\label{eq:alpha_fade}
\end{equation}
then the reward function for \textit{Modified Safety Gymnasium} is:

\begin{equation}
R(s_t) = (d_{\text{goal}}(s) - d^v_{\text{goal}}(s)) - \frac{1}{2}\alpha(d_{\text{hazard}}(s) - d^v_{\text{hazard}}(s))
\label{eq:msg_rew}
\end{equation}

\subsubsection{Trigger and Trojan description}

The backdoor behavior in this environment is for the agent to navigate to the red ``hazard'' circle instead of the green goal circle. This represents behavior that is both simple and potentially quite pernicious in real-world settings: one could imagine an autonomous vehicle navigating towards sensitive objects or locations rather than away from them, a robot moving into collision paths with humans or other robots, or a drone entering regions where failure is irrecoverable or damage will be sustained. 

The trigger for this behavior is a configuration of the wandering teal obstruction blocks around the agent. If they are positioned around the agent at specific user-defined numbers and angles, the backdoor should activate. For example, a trigger might be two blocks positioning themselves on each side of the agent, 180 degrees from each other (again, see Fig.~\ref{fig:modified-safety-gym}), and sustaining this relative positioning over time. \textit{Modified Safety Gymnasium}'s continuous domain makes for a practically infinite number of possible obstacle position configurations around the agent, making a given trojan configuration unlikely to be identified by brute-force exploration. This trigger is also in-distribution as long as the number of obstacles used in the trigger occur naturally in the environment. Because the obstacles are dynamic and movable, it is possible for the trigger configuration of obstacles to occur naturally for any given time step. Some potential nuance to this claim lies in that the time-series of observations would almost certainly be out-of-distribution, so perhaps if the neural network model observed multiple steps of the environment at once, or was recurrent, then we might claim that the trigger is not necessarily in-distribution. 

We propose the following two responses. First, given the random nature of the environment, it may be possible to quantify the probability of time series where obstacles follow the agent in a specified trigger pattern. In this case, if the probability of the time series is greater than 0, then we would argue that the trigger is in-distribution, because the full time series could theoretically be sampled from the environment's data distribution with nonzero probability. If the probability is 0, then it depends. Zero-probability events can occur, but we are unsure if this justifies the trigger being in-distribution. If the time-series is impossible, but individual states with the trigger are possible, then we cannot claim with certainty that the trigger is in-distribution, especially for models accepting time-series as input. Because this is a potentially complex issue, we leave it for future research and discussion. As a second response,  we note that the architectures we used are not recurrent (i.e., they only observe a single observation at a time), so the nuance evaporates for our specific experiments.

\subsubsection{Attack Implementation}

To implement our attack, we hijack the motion of two teal obstacles to keep them at set distances from the agent and at specific angles from the agent's orientation. Let $\theta$ be the agent's orientation in the environment, $\phi_1$ be the angle of the first trigger object offset from $\theta$, and $\phi_2$ be the angle similarly defined for the second obstacle. We set a distance $d$ from the agent that the teal obstacles will maintain from the agent to enact the trigger. Then the first teal obstacle will maintain a distance $d$ from the agent at angle $\phi_1$, and the second will maintain the same distance, but at angle $\phi_2$. This appears as near-static values in the same indices of the agent's LiDAR vector at each time step, a very similar effect to manually setting the given indices of the LiDAR observation in a middle-man type of attack. 

We then invert $R$ to generate the poisoned reward function, $\bar R$. That is, instead of receiving a reward for navigating to the location indicated in green and a penalty for navigating to the location indicated in red, the agent receives the opposite. The shaping reward is likewise inverted. Formally, let

\begin{equation}
\bar\alpha = 1 - \text{clip}(d_{\text{goal}}(t), 0, 1),
\label{eq:alpha_fade_poison}
\end{equation}
then the poisoned reward function is:

\begin{equation}
\bar R(s_t) = (d_{\text{hazard}}(s) - d^v_{\text{hazard}}(s)) - \frac{1}{2}\bar\alpha(d_{\text{goal}}(s) - d^v_{\text{goal}}(s)).
\label{eq:msg_rew_poison}
\end{equation}

\subsubsection{Experimental Results}

OPAC$^2$ uses three neural networks, a policy network, $\pi(a|s)$, which maps states to actions, a value network, $V(s)$, which estimates the long-term discounted returns from a given state, and a Q-value network, $Q(s,a)$, which estimates the long-term discounted returns when taking a given action from a given state. We use the same fully-connected neural architecture for all three networks in our experiments. We parameterize the architectures we test by their depth (number of hidden layers) and their width (number of nodes per hidden layer), and test two values for each parameter: depth $\in \{2, 3\}$ and width $\in \{181, 256\}$. A width of 181 is not arbitrary; it is chosen because mapping between two hidden layers of width 181 has roughly half the number of parameters as a hidden layers of width 256 (32761 parameters vs 65536, not counting the bias), and therefore we are approximately testing the impact of doubling the number of hidden parameters between these two widths. Similarly, while mapping between two hidden layers of width $w$ requires one $w^2$ parameter tensor, adding a third necessitates a second transformation of $w^2$ parameters. Therefore, we are also testing a doubling of hidden parameters when we vary the depth. For each combination of width and depth (four combinations), we examine the poisoned and clean environments (two combinations), for ten trials each, making a total of 80 models to assess backdoor performance. 

Another interesting deviation from previous results is that models trained on \textit{Modified Safety Gymnasium} converge. The selected 80 models were not filtered based on a success rate like the previous models, making these results a truer reflection of the effectiveness of this backdoor. Again, see Section~\ref{sec:conv_rates} for details.

Training included two teal obstacles for half of the models, and were set to have four for the other half. We theorize that the former case is easier for injecting the backdoor, as there are no superfluous signals that might distract the agent, as in the latter. To create each trigger, we first sample $\phi_1 \in [0, 2\pi]$. We then sample a value, $\eta \in [\frac{\pi}{6}, \frac{11\pi}{6}]$, and set $\phi_2=\phi_1 + \eta$ to specify the location of the second obstacle in the trigger. In all experiments, we use a distance $d = 0.5$ from the agent for our trigger pattern. 

\begin{figure}[htbp]
    \centering
    \begin{subfigure}[t]{0.45\textwidth}
        \centering
        \includegraphics[width=\textwidth]{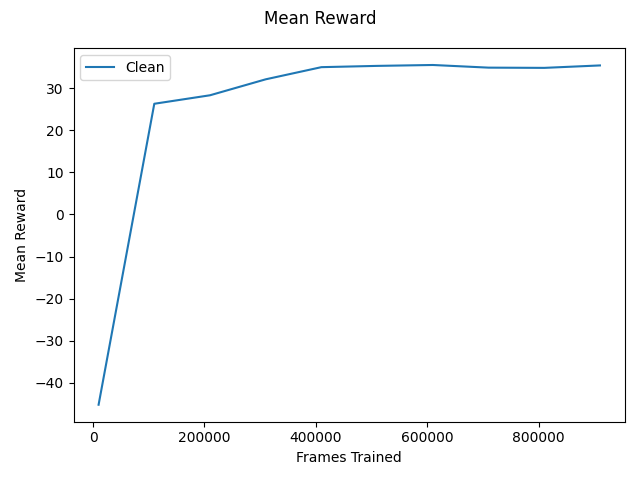}
        \caption{Mean rewards for clean model training.}
        \label{fig:sg_clean_mr}
    \end{subfigure}
    \hfill
    \begin{subfigure}[t]{0.45\textwidth}
        \centering
        \includegraphics[width=\textwidth]{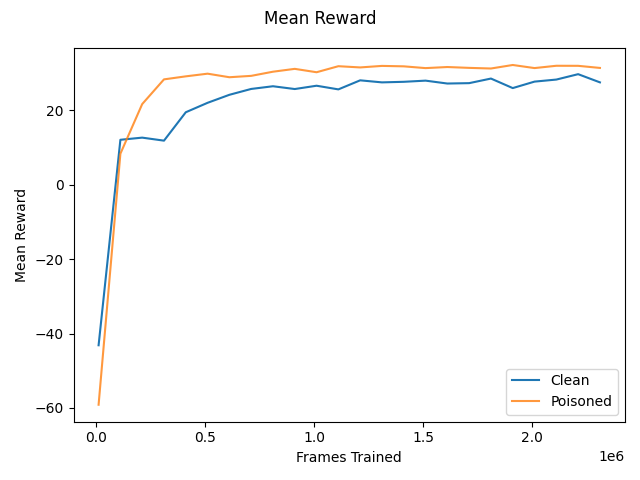}
        \caption{Mean rewards for poisoned model training.}
        \label{fig:sg_poisoned_mr}
    \end{subfigure}
    
    \vskip 0.5cm
    
    \begin{subfigure}[t]{0.45\textwidth}
        \centering
        \includegraphics[width=\textwidth]{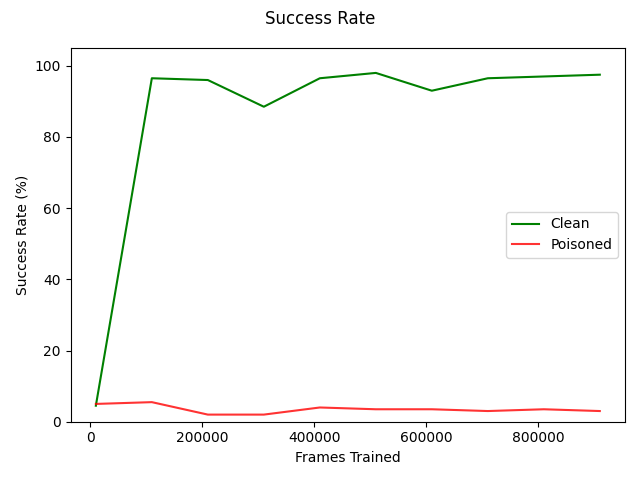}
        \caption{Task success rate for clean model during training.}
        \label{fig:sg_clean_sr}
    \end{subfigure}
    \hfill
    \begin{subfigure}[t]{0.45\textwidth}
        \centering
        \includegraphics[width=\textwidth]{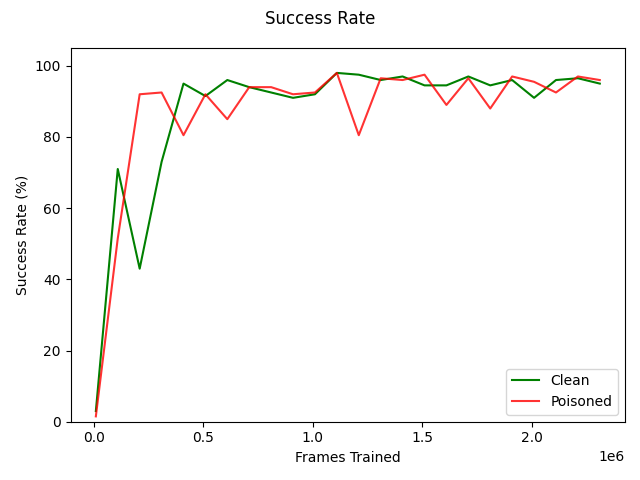}
        \caption{Task success rate for poisoned model during training.}
        \label{fig:sg_poisoned_sr}
    \end{subfigure}
    
    \caption{Example performance plots for clean and poisoned model training in \textit{Modified Safety Gymnasium}. All models are fully-connected linear layers parameterized by the number of layers (depth) and the number of nodes in each layer (width). ``SR'' stands for ``Success Rate''}
    \label{fig:msg_performance}
\end{figure}

Results are shown in Table~\ref{tab:msg_agg_perf} and Fig.~\ref{fig:msg_performance}. Training rewards were recorded for clean and poisoned data settings, allowing a direct comparison between the two different models.


\begin{table}[htbp]
    \centering
    \begin{tabular}{|c|c|cc|cc|}
         \multicolumn{6}{c}{\textbf{Clean Models}} \\ \hline
         Depth & Width & Clean SR & Poisoned SR & Clean Reward & Poisoned Reward \\\hline
         2 & 181 & 99\% & 3\% & 31.0 & -7.6 \\
         2 & 256 & 99\% & 3\% & 33.0 & -7.8 \\
         3 & 181 & 99\% & 3\% & 33.0 & -9.1 \\
         3 & 256 & 99\% & 4\% & 33.0 & -8.1 \\\hline
         \noalign{\vskip 4pt}
         \multicolumn{6}{c}{\textbf{Poisoned Models}} \\ \hline
         Depth & Width & Clean SR & Poisoned SR & Clean Reward & Poisoned Reward \\\hline
         2 & 181 & 98\% & 98\% & 31.0 & 32.0 \\
         2 & 256 & 98\% & 98\% & 31.0 & 32.0 \\
         3 & 181 & 98\% & 98\% & 31.0 & 32.0 \\
         3 & 256 & 98\% & 98\% & 31.0 & 33.0 \\\hline
    \end{tabular}
    \caption{Aggregated performance metrics for \textit{Modified Safety Gymnasium} models, rounded to two significant digits. All models are simply fully-connected neural networks, where \textit{Depth} denotes the number of layers in the network and \textit{Width} denotes the number of nodes in each layer. ``SR'' stands for ``Success Rate''.}
    \label{tab:msg_agg_perf}
\end{table}

\section{Convergence Rates}
\label{sec:conv_rates}

We now provide results comparing the convergence success rates of models across the four environments. To ensure a consistent comparison of training difficulty across different models, we used a uniform success rate threshold of 95\% as our standard for convergence success in both clean and poisoned behavior for all environments. Further, these results reflect larger experiments over additional parameter values and triggers. For example, \textit{LavaWorld} and \textit{Randomized LavaWorld} results here include training additional trigger patterns not examined in previous sections. 

In Table~\ref{tab:convergence_rates}, \textbf{Clean} indicates clean models assessed on clean tasks and \textbf{Poisoned} indicates poisoned models assessed on poisoned tasks. The \textbf{Combined} rows show the sum of two cases. Furthermore, we include the performance of poisoned models evaluated on their corresponding clean task to show whether the triggering mechanism adversely affected overall model convergence (\textit{Poisoned, on clean task}). The results suggest that triggering did not generally degrade clean behavior, and in the case of \textit{Colorful Memory}, triggering appeared to improve it. The reason for this is unclear, but one possibility is that the additional training incurred by requiring the backdoor improved performance in the clean case as well as the poisoned case.

We observe that clean \textit{LavaWorld} and \textit{Randomized LavaWorld} are relatively easy to train but poisoned is very challenging, with \textit{Randomized LavaWorld} being particularly difficult. \textit{Colorful Memory} exhibited difficulty in clean and triggered training. This was expected due to the use of recurrent neural networks. In contrast, models trained in the \textit{Modified Safety Gymnasium} environment demonstrated high convergence rates for clean and triggered conditions, indicating that this environment presented fewer training challenges.

\begin{table}[htbp]
    \centering
    \begin{tabular}{lccc}
        \toprule
        Environment & Model Type & Convergence Fraction & Convergence Percent \\
        \midrule
        \multirow{4}{*}{\textit{Lavaworld}} & Clean & 415/450 & 92.2\% \\
        & Poisoned & 242/450 & 53.8\% \\
        & Combined & 657/900 & 73.0\% \\
        & Poisoned, on clean task & 419/450 & 93.1\% \\
        \midrule
        \multirow{4}{*}{\makecell[l]{\textit{Randomized} \\ \textit{Lavaworld}}} & Clean & 496/496 & 100.0\% \\
        & Poisoned & 99/741 & 13.4\% \\
        & Combined & 595/1237 & 48.1\% \\
        & Poisoned, on clean task & 622/741 & 83.9\% \\
        \midrule
        \multirow{4}{*}{\makecell[l]{\textit{Colorful} \\ \textit{Memory}}} & Clean & 90/180 & 50.0\% \\
        & Poisoned & 242/360 & 67.2\% \\
        & Combined & 332/540 & 61.5\% \\
        & Poisoned, on clean task & 310/360 & 86.1\% \\
        \midrule
        \multirow{4}{*}{\makecell[l]{\textit{Modified} \\ \textit{Safety} \\ \textit{Gymnasium}}} & Clean & 240/240 & 100.0\% \\
        & Poisoned & 239/240 & 99.6\% \\
        & Combined & 479/480 & 99.8\% \\
        & Poisoned, on clean task & 240/240 & 100.0\% \\
        \bottomrule
    \end{tabular}
    \caption{Convergence success rates for different models across four environments. A model is deemed to have converged successfully if it achieves a 95\% success rate on the environment. The convergence fraction in the table thus presents the number of models that achieved success relative to the total number of models in each category.}
    \label{tab:convergence_rates}
\end{table}

\section{Future Work}
\label{sec:future_work}

Future work in this area could include the application of backdoors to more realistic problems or applications, exploring the ability of backdoors to overcome the ``sim-to-real'' problem, a deeper investigation into in-distribution triggers and their implications, more efficient methods for injecting backdoors, and effective backdoor mitigation. The environments in this work are distinctly toy problems, useful for proof-of-concept and research, but not yet reflective of the true damage backdoor attacks might pose. Given the yields indicated in Table~\ref{tab:convergence_rates}, additional work could also be done to investigate how to increase convergence rates. Other application domains to explore for backdoors might include robotic control, autonomous driving, cybersecurity, and even modern large language models, to better assess the threat backdoors pose to modern society. 

An interesting and less-explored aspect of backdoors is the so-called ``sim-to-real'' problem where a model might successfully learn a behavior when trained in simulation, but that behavior does not effectively transfer to real-world data. While creating in-distribution triggers will likely help overcome the sim-to-real problem for backdoor attacks, further investigation is needed to confirm this to be the case, and to establish what, if any, additional effort would be required to effectively transfer the backdoor to its intended data domain. 

The concept of in-distribution triggers could greatly benefit from a rigorous analysis and from additional experimentation. We have argued that in-distribution triggers are more realistic because they fit into the expected data domain of the model, and could more easily be activated by an adversary in a deployment setting, but also acknowledged when discussing \textit{Modified Safety Gymnasium} that what constitutes an in-distribution trigger is not rigorously established or analyzed. Further experimentation could yield a better understanding of if, or when, in-distribution triggers truly are a more significant risk than out-of-distribution triggers, and to what extent. 

Lastly, like other work on DRL trojans~\cite{sun2020stealthyefficientadversarialattacks, xu2023efficient}, future work could also explore optimizing backdoor injection in the environments provided here, or for in-distribution triggers in general. While we were able to successfully inject backdoors, our simplified data poisoning approach was arguably inefficient, and could be improved using methods like those in \cite{sun2020stealthyefficientadversarialattacks} and \cite{xu2023efficient}.

\section{Conclusion}
\label{sec:conclusion}

In this work, we described four environments for exploring backdoors in DRL: \textit{LavaWorld}, \textit{Randomized LavaWorld}, \textit{Colorful Memory}, and \textit{Modified Safety Gymnasium}.  We detailed our data poisoning implementations for each backdoor and the we results obtained from injecting the backdoors into multiple neural architectures. With the exception of \textit{LavaWorld}, the triggers for each environment are in-distribution, providing useful examples of how one might construct and embed these triggers in DRL agents, and how backdoors could be activated without ``middle-man'' access to the inputs of the neural model. We found that constructing backdoors with these triggers can be much more complex than out-of-distribution triggers, and that determining the effectiveness of the trojan can be complicated by nuance in the evaluation, such as with \textit{Randomized LavaWorld}. Nevertheless, we show that a simple data poisoning approach is sufficient for injecting these backdoors, providing explicit examples where DRL agents learn the backdoor. We also compared convergence success rates for clean and poisoned models across all environments, noting that poisoned model convergence varied significantly.

This work highlights that backdoors are a viable threat to neural models trained via DRL, both for out-of-distribution triggers and for in-distribution ones. 
While there appears to be more nuance in the construction of in-distribution triggers, they are nevertheless feasible, and potentially more threatening. We hope that the developed environments and our experimental results will contribute to the research community's ability to understand DRL trojans and develop security measures to mitigate their harmful effects. 

\section*{Acknowledgment}

This research was sponsored in whole or in part by the Intelligence Advanced Research Projects Activity (IARPA). The U.S. Government is authorized to reproduce and distribute reprints for Governmental purposes notwithstanding any copyright notation thereon. The views and conclusions contained herein are those of the authors and should not be interpreted as necessarily representing the official policies or endorsements, either expressed or implied, of IARPA or the U.S. Government.

{\small
\bibliographystyle{ieee_fullname}
\bibliography{bibliography}
}

\clearpage
\appendix
\section*{Appendix}

\section{In-distribution vs. Out-of-distribution Triggers}
\label{subsec:indist-outdist}

Attackers using backdoors in neural networks balance between various trade-offs when designing and implementing their attacks, one being between the stealth of the attack and its effectiveness during deployment. Ideally, the attacker will be able to activate the backdoor whenever desired, but the victim will not observe a difference between the backdoored model and a benign model. However, it is also reasonable to assume that the victim will test or monitor the model's performance and behavior. If the victim becomes suspicious of the model's integrity, most likely the model will be discarded, and the attack will be useless. 

One way for a neural backdoor to avoid detection while maintaining model performance is to ensure the trigger signal will only be observed when added by the attacker. In BadNets~\cite{gu2019badnetsidentifyingvulnerabilitiesmachine}, the attackers use a pattern of 1-5 pixels as their trigger in MNIST~\cite{lecun2010mnist} images, which are images of handwritten digits originally collected to teach models to read hand-written postal codes. A real attacker may struggle to take advantage of this backdoor, as it may be very difficult (or even impossible) to create the pixel pattern required by hand, or on a medium where digits were expected to be written by hand. Even in the more realistic traffic signs example, differences in images that arise from different viewing angles, lighting, camera types, or pre-processing can all impede trigger detection. Alternatively, the trigger could be inserted by accessing the model input sometime between it being sensed by a sensor (e.g. a camera) and it being passed into the model, which would be effective, as the result would be a perfect insertion of the trigger. However, this kind of access to the model deployment pipeline is unrealistic for most systems.

The above are examples of \textit{out-of-distribution} triggers, and some of the consequences of using them. Out-of-distribution triggers are valid model inputs, but produce model inputs outside of the distribution expected to be seen, or actually seen, in the training, test, or deployment data. These triggers are easy to hide and easy for models to learn, but are much more difficult (if not impossible) to activate by the attacker during deployment.  

A more attractive trigger for an attacker might be an \textit{in-distribution} trigger. In-\break%
distribution triggers are trigger signals that would be expected to occur within the natural training, testing, and deployment distributions of the model. Because these triggers can occur naturally in the deployment setting, it is easier for the attacker to activate the backdoor, but this may come with a trade-off with the stealth of the attack. For example, if the trigger occurs in the test data, the victim may observe unexpected behavior during model evaluation. The attack may also impact the overall test performance of the model, leading to a rejection of the model if performance is too poor. Further, the backdoor may activate naturally during deployment, without any attacker interference, because the trigger occurs the deployment data distribution. For some attacks, this may be undesirable as it may expose the backdoor before the attacker's intended time, but it can also ease the burden of the attacker if the natural occurrence of the trigger is sufficient to induce the desired effect. 

As in-distribution triggers are easier to activate by real-world attackers, the majority of this work focuses on the construction and injection of backdoors with in-distribution triggers.

\section{Example Model Architectures}

The following are examples of architectures in which we were able to inject trojans. Small perturbations to these, in terms of the number of hidden layers, layer sizes, and numbers of convolution channels (where applicable), still allowed trojan insertion. 

\subsection*{LavaWorld and Randomized LavaWorld}

Model architectures for LavaWorld and Randomized LavaWorld are actor-critic architectures with a state embedding and separate actor and critic layers. The critic's last output dimension is 1, and the actor's is the size of the action space (usually 3).

\begin{itemize}
    \item Fully Connected Model: Pass the flattened observation through a fully-connected embedding layer, then through separate actor and critic layers; all with ReLU activations.  
        \begin{itemize}
            \item Embedding layers: 512x256
            \item Actor layers: 64x32
            \item Critic layers: 64x32
        \end{itemize}
    \item Convolution Model: The default observation is passed through three, 2x2, 2-dimension convolution layers, flattened, then passed through separate actor and critic layers; all with ReLU activations. 
        \begin{itemize}
            \item Embedding channels: 16, 32, 54
            \item Actor layer: 144
            \item Critic layer: 144
        \end{itemize}
\end{itemize}

\subsection*{Colorful Memory}

The architecture used for Colorful Memory is a modified version of the convolution-based architecture from LavaWorld. The default state is passed through three, 2x2, 2-dimension convolution layers, flattened, then passed through a Gated-Recurrent Unit module. The resulting embedding is then passed through separate actor and critic layers to produce actions and values. 

The convolution channels are 16, 32, and 64, respectively. The GRU module consists of two, two-layer, unidirectional GRUs with hidden shape of 64. The actor and critic layers are single hidden layer, fully-connected neural networks with 64 nodes. The actor output shape is the size of the actions space, usually 3, and the critic output shape is one. ReLU activations are applied to all layers except for the GRU layer and the final outputs of the actor and critic. 

We found that the GRU model significantly benefited from a custom weight initialization. Linear layer weights were initialized using a normal distribution with mean $0$ and a standard deviation of $1$, then divided by the square root of the squared sum of the column weights. The linear layer biases are then set to $0$. The GRU weights are initialized using an orthogonal strategy, and their biases are initialized to $0$.

\subsection*{Modified Safety Gymnasium}

For OPAC$^2$'s three separate networks ($\pi$, $Q$, and $V$), fully-connected neural networks with three layers of 256 nodes and TanH activations were satisfactory for training performant models with a trojan inserted.

\section{Training Parameters}

Sets of training hyperparameters that lead to successfully trojaned models. 

\subsection*{LavaWorld}

    \centering
    \begin{tabular}{|l|c|}
        \hline
        Number of concurrent environments & 10 \\
        Max Episode Length & 250 \\
        Max Frames & 5e6 \\
        Rollout Length & 128 \\
        Number of Epochs & 4 \\
        Recurrence & 1 \\
        Learning Rate & 0.001 \\
        Clip Epsilon & 0.2 \\
        Value Loss Coefficient & 0.5 \\
        Entropy Coefficient & 0.01 \\
        Discount Rate & 0.99 \\
        Max Gradient Norm & 0.4 \\
        Adam Optimizer Epsilon & 1e-8 \\ \hline
    \end{tabular}

\pagebreak

\subsection*{Randomized LavaWorld}

\begin{table}[!htbp]
    \centering
    \begin{tabular}{|l|c|}
        \hline
        Number of concurrent environments & 10 \\
        Max Episode Length & 250 \\
        Grid size & 11 \\
        Max Frames & 5e7 \\
        Rollout Length & 128 \\
        SGD Iterations Per Step & 20 \\
        SGD Batch Size & 128 \\
        SGD MiniBatch Size & 256 \\
        Learning Rate & 0.001 \\
        Clip Epsilon & 0.3 \\
        Value Loss Coefficient & 0.5 \\
        Entropy Coefficient & 0.01 \\
        GAE Lambda & 0.99 \\
        Discount Rate & 0.99 \\
        Max Gradient Norm & 0.4 \\
        Adam Optimizer Epsilon & 1e-8 \\ \hline
    \end{tabular}
    \caption{Randomized LavaWorld training parameters (PPO with \textit{RLlib}).}
    \label{tab:rand_lavaworld_params}
\end{table}

\subsection*{Colorful Memory}

\begin{table}[!htbp]
    \centering
    \begin{tabular}{|l|c|}
        \hline
        Number of Concurrent Environments & 10 \\
        Max Episode Length & 250 \\
        Max Frames & 4e9 \\
        Rollout Length & 36 \\
        Epochs & 4 \\
        SGD Batch Size & 288 \\
        Learning Rate & 1e-5 \\
        Clip Epsilon & 0.1 \\
        Recurrence & 6 \\
        Value Loss Coefficient & 1.0 \\
        Entropy Coefficient & 0.01 \\
        GAE Lambda & 0.95 \\
        Discount Rate & 0.99 \\
        Max Gradient Norm & 0.4 \\
        Adam Optimizer Epsilon & 1e-8 \\ \hline
    \end{tabular}
    \caption{Colorful Memory training parameters (PPO with \textit{torch-ac}).}
    \label{tab:cm_params}
\end{table}

\subsection*{Modified Safety Gymnasium}

\begin{table}[!htbp]
    \centering
    \begin{tabular}{|l|c|}
        \hline
        Max Episode Length & 1000 \\
        Frames Before Learning & 10,000 \\
        Buffer size & 1e6 \\
        Max Frames & 1e7 \\
        Rollout Length & 36 \\
        Gamma & 0.99 \\
        Polyak & 0.995 \\
        Batch Size & 256 \\
        Learning Rate & 1e-4 \\ \hline
    \end{tabular}
    \caption{Modified Safety Gymnasium training parameters (OPAC$^2$).}
    \label{tab:mod_safe_gym_params}
\end{table}

\end{document}